%% file: moonshine_paper.tex
\documentclass[nohyperref]{article}

\usepackage{microtype}
\usepackage{graphicx}
\usepackage{subfigure}
\usepackage{booktabs} 
\usepackage[dvipsnames]{xcolor}
\usepackage{soul}
\usepackage{dblfloatfix}
\usepackage{import}

\usepackage{hyperref}
\usepackage{multirow}
\usepackage{tabularx}
\usepackage{colortbl}
\usepackage{hhline}

\newcolumntype{C}{>{\centering\arraybackslash}X}

\usepackage[accepted]{icml2022}

\usepackage{amsmath}
\usepackage{amssymb}
\usepackage{mathtools}
\usepackage{amsthm}

\theoremstyle{plain}

\theoremstyle{definition}

\theoremstyle{remark}

\usepackage[textsize=tiny]{todonotes}

\icmltitlerunning{Flavors of Moonshine: Tiny Specialized ASR Models for Edge Devices}
\begin{document}

\twocolumn[
  \icmltitle{Flavors of Moonshine: Tiny Specialized ASR Models for Edge Devices}



\icmlsetsymbol{equal}{*}

\begin{icmlauthorlist}
\icmlauthor{Evan King}{equal}
\icmlauthor{Adam Sabra}{equal}
\icmlauthor{Manjunath Kudlur}{equal}
\icmlauthor{James Wang}{}
\icmlauthor{Pete Warden}{}
\end{icmlauthorlist}

\begin{center}
\href{https://moonshine.ai}{Moonshine AI}
\end{center}
\icmlcorrespondingauthor{Pete Warden}{pete@moonshine.ai}

\icmlkeywords{Speech Recognition, Machine Learning, TinyML}

\vskip 0.3in
]



\printAffiliationsAndNotice{$^{*}$Equal contribution} 

\begin{abstract}
We present the Flavors of Moonshine, a suite of tiny automatic speech recognition (ASR) models specialized for a range of underrepresented languages. Prevailing wisdom suggests that multilingual ASR models outperform monolingual counterparts by exploiting cross-lingual phonetic similarities. We challenge this assumption, showing that for sufficiently small models (27M parameters), training monolingual systems on a carefully balanced mix of high-quality human-labeled, pseudo-labeled, and synthetic data yields substantially superior performance. On average, our models achieve error rates 48\% lower than the comparably sized Whisper Tiny model, outperform the 9x larger Whisper Small model, and in most cases match or outperform the 28x larger Whisper Medium model. These results advance the state of the art for models of this size, enabling accurate on-device ASR for languages that previously had limited support. We release Arabic, Chinese, Japanese, Korean, Ukrainian, and Vietnamese Moonshine models under a permissive open-source license.
\end{abstract}

\section{Introduction}
\label{sec:introduction}
\import{sections}{introduction.tex}

\section{Approach}
\label{sec:approach}
\import{sections}{approach.tex}

\section{Evaluations}
\label{sec:evaluation}
\import{sections}{evaluation.tex}

\section{Discussion \& Conclusion}
\label{sec:conclusion}
\import{sections}{conclusion.tex}


\nocite{*}
\bibliography{moonshine_paper}
\bibliographystyle{icml2022}

\appendix
\onecolumn
\import{sections}{appendix.tex}

\end{document}

%% file: sections/introduction.tex
Automatic Speech Recognition (ASR) has seen growing interest in recent years, driven by new opportunities for speech-driven human-machine interaction~\cite{namvarpour2025art}. Low latency ASR models that are small enough to deploy on-device can empower a new generation of voice-driven applications, from real-time translation devices to intelligent conversational interfaces in automobiles~\cite{rege2024talking} and smart appliances~\cite{king2025teaching}. While resource-intensive ASR models like Whisper Large~\cite{radford2023robust} and NVIDIA Canary~\cite{puvvada2024less}, or multimodal models like Phi-4~\cite{abouelenin2025phi} can be deployed in the cloud, they incur substantial infrastructure costs, rely on internet connectivity, and raise user privacy concerns. These limitations have spurred the development of ASR models that are compact enough to run efficiently on the edge~\cite{jeffries2024moonshine}.

To date, most lightweight ASR research has focused on English or on multilingual models that underperform on non-English languages. Whisper Tiny is a multilingual model that is small enough to run on-device, and provides a good example: while it achieves a strong 12\% error rate on English, the model is less-than-stellar for Vietnamese ASR, where it has a 60\% error rate~\cite{radford2023robust}. In other words, while Whisper Tiny \emph{technically} supports Vietnamese, its usability in real-world on-device applications is limited. This performance gap motivates the development of tiny ASR models that better support non-English languages.

Prior work suggests that multilingual models can leverage cross-lingual similarities, allowing knowledge from high-resource languages to transfer and improve recognition in lower-resource ones~\cite{cho2018multilingual,toshniwal2018multilingual,ardila2019common,pratap2020massively}. However, training models that are lightweight enough for edge devices requires reducing the number of learnable parameters, meaning that convergence on a truly high-performance multilingual model is challenging. In this paper, we focus our efforts on training lightweight (27M parameter) \emph{monolingual} models for edge devices, exploiting efficiencies in training data that we achieve from a combination of open human-labeled datasets, high-quality pseudo-labels, and synthetic utterances. We choose six languages---Arabic, Chinese, Japanese, Korean, Ukrainian, and Vietnamese---with varying levels of training data availability. Our models achieve word error and character error rates (WER/CER) that are on-average 48\% lower than the comparably sized Whisper Tiny model, and are in most cases on-par or better than the 28x larger Whisper Medium model. Additionally, by exploiting the architectural benefits of the Moonshine model architecture, our models run between 5x-15x faster than Whisper in on-device applications.

In summary, this paper introduces the following:

\begin{itemize}
    \item We present new Moonshine ASR models for Arabic, Chinese, Japanese, Korean, Ukrainian, and Vietnamese. We show that the models achieve an average of 48\% lower WER/CER than Whisper Tiny, making them significantly better-suited to edge ASR applications.
    \item We show that in all cases, the models outperform the 9x larger Whisper Small model. We also show that in most cases, the models are on par or 5-10\% better than Whisper Medium, which is 28x larger.
    \item We release the models under a permissive open-source license.
\end{itemize}

The remainder of this paper is structured as follows. Section~\ref{sec:approach} describes our approach to training data collection, preprocessing, and model training. Section~\ref{sec:evaluation} summarizes the results of our evaluations. Section~\ref{sec:conclusion} concludes the paper. Appendices~\ref{sec:public-datasets}, \ref{sec:normalization}, and~\ref{sec:exhaustive} include information about public datasets, evaluation procedure, and detailed results.

\input{figures/tables/data_composition.tex}

%% file: figures/tables/data_composition.tex
\begin{table}[ht]
    \centering
    \begin{tabular}{lrrr|r}
    \toprule
    \textbf{Language}   &   \textbf{Open} &   \textbf{Internal} &   \textbf{Synthetic} &   \textbf{Total} \\
    \midrule
    Arabic     &    4.6 &            10.0 &         0.9 &    15.5 \\
    Chinese    &   50.9 &            19.0 &         0.0 &    69.8 \\
    Japanese   &   36.9 &            17.0 &         0.0 &    53.9 \\
    Korean     &   27.6 &            44.4 &         0.0 &    72.0 \\
    Ukrainian  &    1.7 &            12.9 &         5.1 &    19.6 \\
    Vietnamese &    8.4 &            85.8 &         0.0 &    94.2 \\
    \bottomrule
    \end{tabular}
    \caption{Training data hours (in thousands) by language and source. We supplement existing publicly-released datasets with internally collected datasets and, in some cases, synthesize utterances for lower-resource languages.}
    \label{tab:hours_by_language}
\end{table}

%% file: sections/approach.tex
This section describes the Moonshine model architecture before detailing our data preparation and model training process.

\subsection{Architecture}

We leverage the Moonshine model architecture~\cite{jeffries2024moonshine}, an encoder-decoder transformer that applies rotary position embeddings (RoPE) in its encoder and decoder layers~\cite{su2024roformer}. Compared to Whisper, Moonshine is especially well-suited for edge applications because its inference cost---in terms of FLOPs and corresponding latency---scales with the duration of the input audio. In contrast, Whisper pads all inputs to 30 seconds, regardless of length, leading to unnecessary computation. We adopt the Moonshine Tiny variant (27M parameters), which is small enough to be deployed in resource-constrained environments. Table~\ref{tab:model-shapes} provides a comparison between the Moonshine Tiny architecture and the similarly-sized Whisper Tiny model.

\input{figures/tables/architecture_comparison.tex}

\subsection{Data collection, preprocessing, \& synthesis}

The amount of data used as input for training a transformer-based ASR model is loosely predictive of model performance, with a minimum of between $10^4$ to $10^5$ hours required for usable results~\cite{radford2023robust}. Some languages are considered ``higher-resource'' than others---that is, there is more recorded audio available that can be used for training. As we are focused on a range of mid to low-resource languages, the availability of data varies for each. With this in mind, we approached data collection, preprocessing, and, in some cases, synthesis, using a three-stage strategy applied to each language:

\begin{enumerate}
    \item \textbf{Aggregate publicly-available ASR datasets.} We leveraged ASR datasets from prior work. This allowed us to establish a baseline for model performance---effectively the best results we could achieve without performing raw data collection and labeling.
    \item \textbf{Collect and pseudo-label publicly-available data.} Because existing public datasets lack the volume needed to train high-performing models, we gathered and pseudo-labeled a large collection of raw audio from publicly available sources such as podcasts and radio streams.
    \item \textbf{Synthesize labeled data from text-only datasets.} In cases where enough raw data was not widely available, we used high-quality text-to-speech models to synthesize diverse utterances from traditionally text-only datasets.
\end{enumerate}

Table~\ref{tab:hours_by_language} provides an overview of the sources and sizes of training datasets for each language. Appendix~\ref{sec:public-datasets} provides a detailed breakdown of the public datasets for each language.

Since we are targeting lower-resource languages, publicly available datasets are insufficient to achieve high performance. Our data collection target was to exceed the amount (in hours) used to train the initial Whisper models, based on the intuitions that (1) monolingual models do not benefit from transfer learning, and thus need larger datasets and (2) model performance loosely scales with dataset size. We therefore collected and labeled a large amount of raw, in-the-wild speech data available on the open internet. Leveraging WhisperX~\cite{bain2022whisperx} modified to run in a custom framework for distributed data processing, we prepared around 173,000 hours of internal datasets across languages. In all cases except Chinese~\footnote{Since Whisper's release, the amount of publicly-available Chinese data has grown substantially, which reduces the need for internal collection.}, our monolingual datasets exceeded the size of those used to train the original Whisper by an order of magnitude.

In the cases of Arabic and Ukrainian, raw data sources on the open internet were insufficient to meet the threshold for dataset hours. To fill the gap, we leveraged text-only datasets for each language to create uniform distributions of utterance lengths, which we then input to high-quality, high-diversity text-to-speech models to generate fully synthetic utterances. Employing style interpolation between speaker embeddings helped us ensure a diverse set of synthetic speakers~\cite{dumoulin2016learned}.

\subsection{Training}

We train all models using a schedule-free AdamW optimizer \cite{defazio2024road} with a learning rate of $2e^{-5}$ for 8 epochs, with a batch size of 32. All models were trained via DDP on 8xH100 GPUs until completion. Training took between 1 to 3 days depending on the amount of data in our language-specific corpus.

%% file: figures/tables/architecture_comparison.tex
{
  \renewcommand{\arraystretch}{1.2}
  \begin{table}[h]
    \small
    \centering
    \begin{tabular}{|c|c|c|}
    \hline
    \textbf{Parameter} & \textbf{Moonshine} & \textbf{Whisper} \\
    \hline \hline
    Dimension & 288 & 384 \\
    \hline
    Encoder layers & 6 & 4 \\
    \hline
    Decoder layers & 6 & 4 \\
    \hline
    Attention heads & 8 & 6 \\
    \hline
    Encoder FFN activation & \multicolumn{2}{|c|}{GELU} \\
    \hline
    Decoder FFN activation & SwiGLU & GELU \\
    \hline
    Parameters (M) & 27.1 & 37.8 \\
    \hline
    FLOPs vs. Whisper Tiny & 0.7x & 1.0x \\
    \hline
    \end{tabular}
    \caption{Tiny model architectures and inference FLOPs}
    \label{tab:model-shapes}
  \end{table}
}

%% file: sections/evaluation.tex
\input{figures/tables/tiny_error_comparison.tex}

\input{figures/tables/all_error_delta.tex}

Our evaluations measure the performance of Moonshine Tiny models against several multilingual Whisper variants of increasing size---Tiny, Base, Small, and Medium. We choose these Whisper variants since they are small enough to be deployed in the same on-device applications we target with Moonshine. We use a beam size of $1$ in evaluations with Whisper, which is functionally equivalent to the greedy decoding used by Moonshine~\footnote{In the original Whisper paper, the authors use a beam size of 5. This produces slightly different results than the ones we report in this paper.}.

Our primary metric is error rate. Both word error rate (WER) and character error rate (CER) are commonly used in evaluating ASR systems; however, there is some subjectivity as to the best metric for certain languages. For completeness, we measure both WER and CER in every evaluation (except Japanese and Chinese, which lack word boundaries) and report the exhaustive results in Appendix~\ref{sec:exhaustive}. The summarized results in this section only report one or the other as is appropriate for the language.

We rely on two multilingual test sets for evaluating every language: Common Voice 17~\cite{ardila2019common} and Fleurs~\cite{conneau2023fleurs}. For some languages, we also evaluate using language-specific test sets. Arabic uses the test sets proposed by the Open Universal Arabic ASR Leaderboard~\cite{wang2024open}, Japanese uses Reazon Speech~\cite{fujimoto2016reazonspeech}, Korean uses Zeroth-Korean~\cite{zeroth-korean_slr40}, and Ukrainian uses Eurospeech~\cite{disco-eth_eurospeech_2025}. 

To ensure consistent evaluations across datasets, models, and languages, we normalize both the reference and predicted transcription prior to calculating the error rate.

\subsection{Results}



\begin{figure*}[t]
  \centering
  \includegraphics[width=\textwidth]{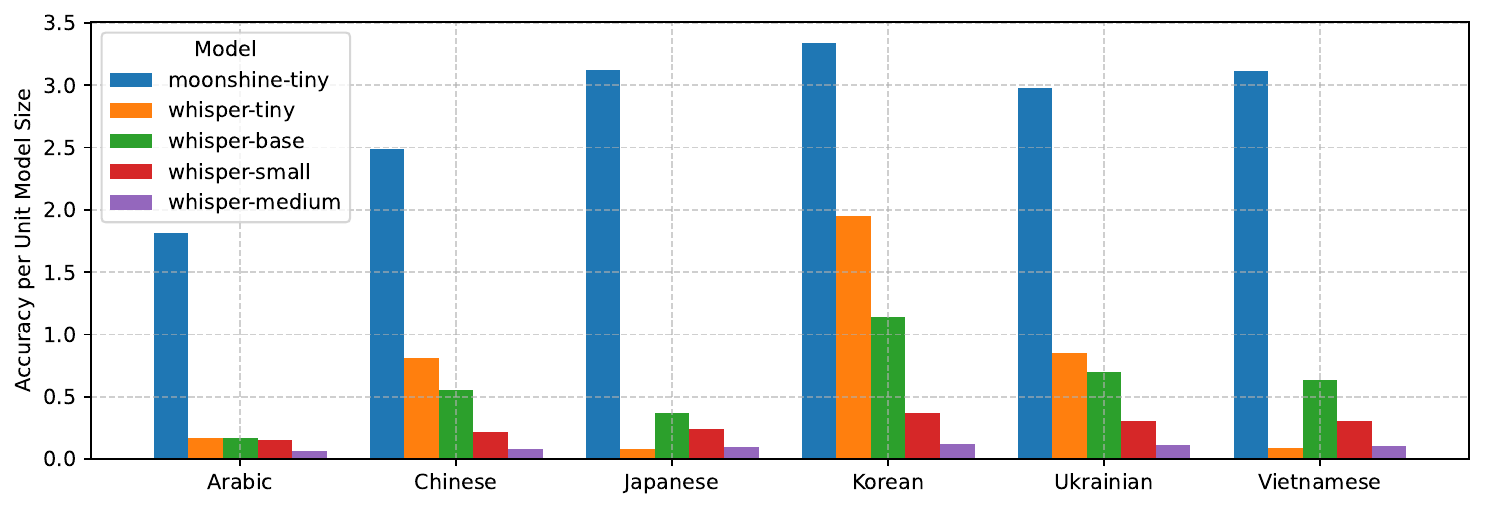}
  \caption{Difference in accuracy per model size between Moonshine and Whisper models. Model size places an upper bound on performance, and some models manage this tradeoff better than others. Moonshine offers a superior tradeoff between performance and size than Whisper models its size and larger.}
  \label{fig:unit-cost}
\end{figure*}

\begin{figure*}[t]
  \centering
  \includegraphics[width=\textwidth]{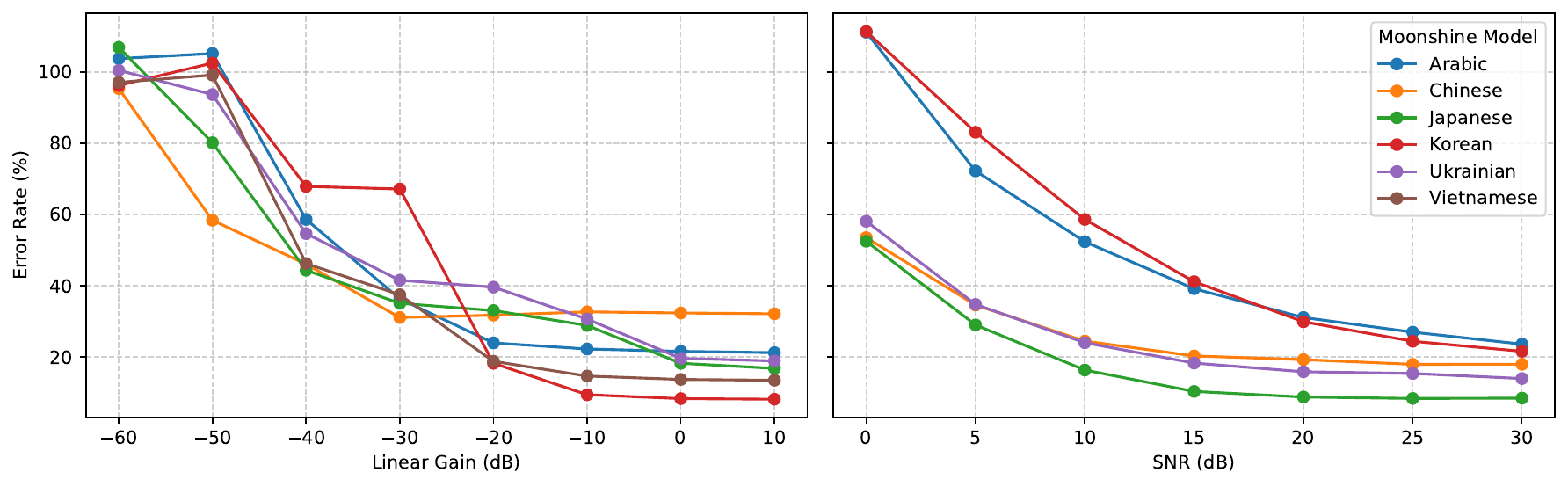}
  \caption{Effect of input audio gain and signal-to-noise ratio (SNR) on Moonshine Tiny error rate.}
  \label{fig:gain-snr}
\end{figure*}

\begin{figure}
  \centering
  \includegraphics[width=\columnwidth]{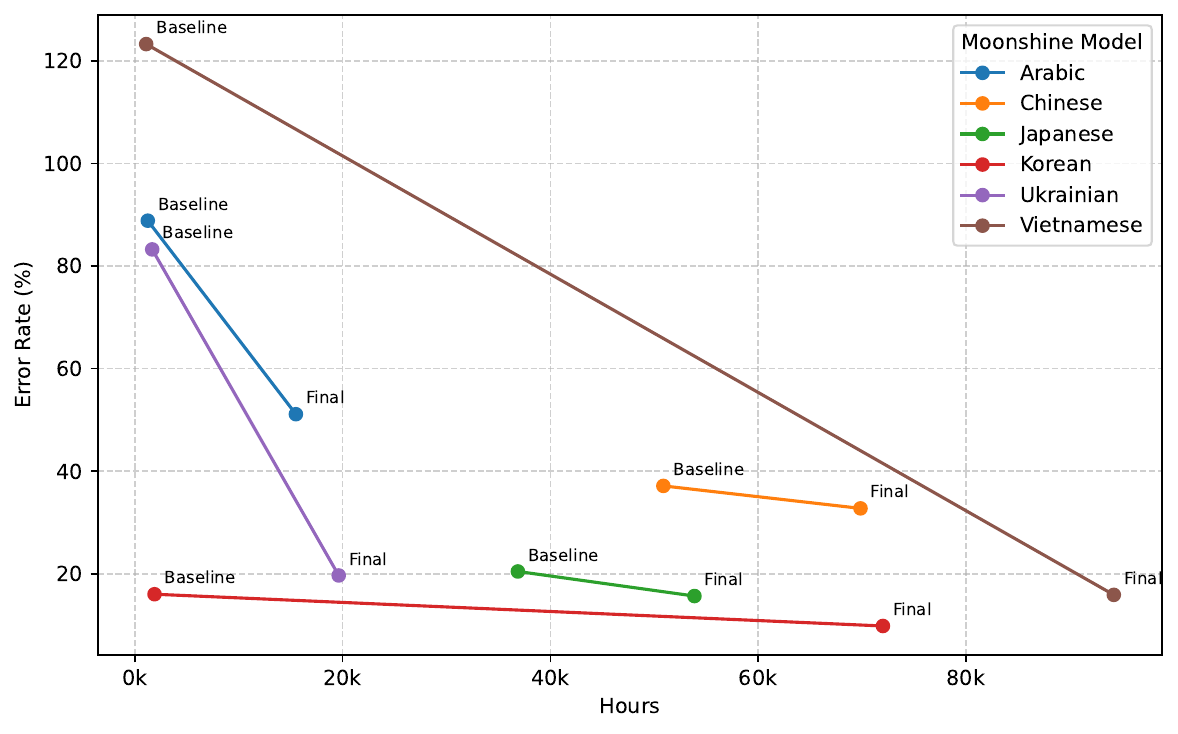}
  \caption{Error rate vs. hours of training data. Model performance scales loosely with training data hours, dependent on data quality.}
  \label{fig:data-error}
\end{figure}

\textbf{Moonshine outperforms Whisper models between 1x and 28x its size.} We first compare Moonshine Tiny with the similarly-sized Whisper Tiny model. Table~\ref{tab:error-rate} shows that the monolingual Moonshine Tiny models significantly outperform Whisper Tiny on all tests. Table~\ref{tab:error-rate-delta} further illustrates the difference in error rate between Moonshine and progressively larger Whisper models (averaged across all evaluations for a language). These results show that Moonshine outperforms the 9x larger Whisper Small model, attaining performance that is on-par or better than the 28x larger Whisper Medium model.

There is an inherent tradeoff between model size and performance: more compute resources allow for larger models, which have a higher upper bound of performance. We compare this tradeoff using a measure of models' ``unit accuracy'', i.e., the inverse of error rate divided by model size $(100 - \text{Error})/(\text{\# of Parameters})$. Figure~\ref{fig:unit-cost} visualizes this tradeoff for Moonshine and Whisper, showing that Moonshine Tiny generally offers a superior tradeoff between compute requirements and accuracy than Whisper models its size and larger.

\textbf{Performance scales loosely with dataset size.} Figure~\ref{fig:data-error} compares two checkpoints of Moonshine Tiny training (``Baseline'' and ``Final'') with increasing hours of training data. Model performance improves as the size of the training set increases, but the rate of increase is not consistent across languages. Vietnamese, for instance, has a large increase in data and a correspondingly large increase in performance; Korean, on the other hand, has an equally large increase in data, but the performance benefits are less significant.

\textbf{Performance degrades gracefully with reduced gain and increased noise.} Figure~\ref{fig:gain-snr} depicts the error rate of each model when evaluated on Fleurs with varying linear gain and signal-to-noise ratio. This allows us to assess the effect of input audio gain (quiet to loud) and audio noise on model performance. Performance trends lower as gain decreases and noise increases, though the models are robust to around -20 dB gain and 20 dB SNR, respectively. Interestingly, Ukrainian performance does not degrade more substantially than other languages despite being trained on a relatively larger amount of clean audio from synthetic speakers. End-to-end applications can leverage methods for input gain normalization and noise suppression to provide added robustness beyond the model's baseline capabilities.

%% file: figures/tables/tiny_error_comparison.tex
\renewcommand{\arraystretch}{1.2}

\begin{table*}[t]
    \centering
        \begin{tabular}{clrrrrrr}
        \cline{2-8}
        & \textbf{Model}   &   \textbf{Arabic} &   \textbf{Chinese} &   \textbf{Japanese} &   \textbf{Korean} &   \textbf{Ukrainian} &   \textbf{Vietnamese} \\
        \hline
        \multirow{2}{*}{CV17} & whisper-tiny     &              88.9 &                66.0 &                96.1 &              37.3 &                 67.1 &                 100.9 \\
        & moonshine-tiny   &              \cellcolor{green!10}36.6   &                 \cellcolor{green!10}36.1 &                \cellcolor{green!10}18.3 &              \cellcolor{green!10}14.9 &                 \cellcolor{green!10}26.1 &                  \cellcolor{green!10}18.8 \\
        \hline
        \hline
        \multirow{2}{*}{Fleurs} & whisper-tiny     &              66.0 &                71.1 &                47.2 &              15.8 &                 63.8 &                  91.9 \\
        & moonshine-tiny   &              \cellcolor{green!10}20.8   &                \cellcolor{green!10}29.4 &                \cellcolor{green!10}17.9 &               \cellcolor{green!10}8.9 &                 \cellcolor{green!10}18.2 &                  \cellcolor{green!10}13.0 \\
        \hline
        \end{tabular}
    \caption{Error rates of Moonshine Tiny and Whisper models across languages. Chinese, Japanese, and Korean results are CER; other languages are WER. Specialized Moonshine Tiny models outperform multilingual Whisper Tiny on all languages.}
    \label{tab:error-rate}
\end{table*}

%% file: figures/tables/all_error_delta.tex
\renewcommand{\arraystretch}{1.2}

\begin{table*}[t]
    \centering
    \begin{tabular}{l|l|llllll}
    \hline
    \textbf{Comparison}   & \textbf{Relative Size} & \textbf{Arabic}           & \textbf{Chinese}          & \textbf{Japanese}         & \textbf{Korean}           & \textbf{Ukrainian}        & \textbf{Vietnamese}       \\
    \hline
    vs. whisper-tiny      & 1.4x & \cellcolor{green!10}-42.4 & \cellcolor{green!10}-35.7 & \cellcolor{green!10}-81.0 & \cellcolor{green!10}-14.1 & \cellcolor{green!10}-47.1 & \cellcolor{green!10}-80.5 \\
    vs. whisper-base      & 2.7x & \cellcolor{green!10}-36.4 & \cellcolor{green!10}-26.4 & \cellcolor{green!10}-57.0 & \cellcolor{green!10}-6.1  & \cellcolor{green!10}-28.9 & \cellcolor{green!10}-36.9 \\
    vs. whisper-small     & 9.0x & \cellcolor{green!10}-12.1 & \cellcolor{green!10}-14.0 & \cellcolor{green!10}-25.3 & \cellcolor{green!10}-0.0  & \cellcolor{green!10}-6.2  & \cellcolor{green!10}-10.5 \\
    vs. whisper-medium    & 28.5x & 1.0                       & \cellcolor{green!10}-7.6  & \cellcolor{green!10}-12.2 & 2.2                       & 3.2                       & \cellcolor{green!10}-2.6  \\
    \hline
    \end{tabular}
    \caption{Error rate difference between Moonshine Tiny and all Whisper models; lower is better. Chinese, Japanese, and Korean results are CER; other languages are WER. Moonshine Tiny outperforms Whisper Small, which is 9x larger; in some cases, Moonshine Tiny outperforms the 28x larger Whisper Medium model.}
    \label{tab:error-rate-delta}
\end{table*}

%% file: sections/conclusion.tex
This section discusses limitations and future work before concluding the paper.

\textbf{Low and ultra-low resource languages.} We aim to further expand our scope to low and ultra-low resource languages, leveraging similar data collection and pseudo-labeling techniques outlined in the paper. More extensive use of synthesis, as well as introduction of data augmentation techniques may be necessary to fill the gaps in raw data available.

\textbf{Language-specific tokenizers.} Moonshine currently relies on GPT-2's multilingual tokenizer, which has a larger vocabulary size than necessary for individual languages. We expect that reducing the vocabulary size for each model will simplify the next-token prediction task, with potential benefits to accuracy and latency.

In summary, we introduce a family of 27M parameter Moonshine Tiny models that match or outperform the 28x larger Whisper Medium model across 6 languages. Our experiments show that small, monolingual, and specialized ASR models are better suited for on-device tasks in underrepresented languages. To contribute to the broader research effort around these tasks, we release the open weights to all the models discussed in the paper with a permissive license.

%% file: sections/appendix.tex
\section{Public Datasets}
\label{sec:public-datasets}
    \begin{table}[ht]
        \centering
        \begin{tabularx}{\columnwidth}{lXr}
            \toprule
            \textbf{Language}   & \textbf{Dataset (Hugging Face)}                                        &   \textbf{Hours} \\
            \midrule
            Arabic     & NeoBoy/arabic-tts-wav-24k~\cite{kulkarni2023clartts, arabictts24k}                      &      12 \\
                 & nadsoft/arabic-98                              &     163 \\
                 & MAdel121/arabic-egy-cleaned                    &      72 \\
                 & Sabri12blm/Arabic-Quran-ASR-dataset            &     360 \\
                 & MohamedRashad/SADA22~\cite{alharbi2024sada}                           &     647 \\
                 & NightPrince/MasriSpeech-Full~\cite{masrispeech_full}                   &    3100 \\
                 & xmodar/commonvoice-12.0-arabic-voice-converted~\cite{ardila2019common} &     148 \\
                 & google/fleurs~\cite{conneau2023fleurs}                                  &     116 \\ \hline
            Chinese    & Hannie0813/NVSpeech170k~\cite{liao2025nvspeech}                        &     573 \\
                & amphion/Emilia-Dataset~\cite{he2024emilia}                         &   50300 \\ \hline
            Japanese   & reazon-research/reazonspeech~\cite{fujimoto2016reazonspeech}                   &   35000 \\
               & kadirnar/Combined-Japanese-TTS                 &     995 \\
               & mozilla-foundation/common\_voice\_17\_0~\cite{ardila2019common}        &     610 \\
               & amphion/Emilia-Dataset~\cite{he2024emilia}                         &     266 \\ \hline
            Korean     & brainer/korean-telemedicine-speech             &    1120 \\
                 & imTak/korean-audio-text-economy                &      39 \\
                 & Junhoee/STT\_Korean\_Dataset                   &     267 \\
                 & JaepaX/korean\_dataset                         &     424 \\
                 & Bingsu/zeroth-korean~\cite{zeroth-korean_slr40}                           &      52 \\
                 & jp1924/GyeongsangSpeech                        &    2546 \\
                 & amphion/Emilia-Dataset~\cite{he2024emilia}                         &    7500 \\
                 & jp1924/KoreanUniversityLectureData             &    4000 \\
                 & jp1924/KoreaSpeech                             &    3630 \\
                 & jp1924/KconfSpeech                             &    2970 \\
                 & jp1924/KrespSpeech                             &    2907 \\
                 & jp1924/JeollaSpeech                            &    2121 \\ \hline
            Ukrainian  & disco-eth/EuroSpeech~\cite{disco-eth_eurospeech_2025}                           &    1287 \\
              & speech-uk/voice-of-america~\cite{smoliakov_2025}                    &     391 \\ \hline
            Vietnamese & NhutP/VietSpeech~\cite{VietSpeech}                               &    1100 \\
             & speechcolab/gigaspeech2~\cite{gigaspeech2}                        &    7324 \\
            \bottomrule
        \end{tabularx}
        \caption{Publicly-available training datasets. We include citations for datasets that have an associated paper, or that have citation instructions on the repo at time of writing.}
        \label{tab:open_hours}
\end{table}

\section{Normalization Steps}
\label{sec:normalization}

This section outlines the normalization steps for each language.

To normalize Arabic, we leveraged the normalization code provided by \cite{chowdhury2021modelruleallmultilingual, wang2024open}. To be specific, we removed punctuation and diacritics, and converted Eastern Arabic numerals to Western Arabic numerals. For Korean, we leveraged the normalizing steps proposed by Zeroth~\footnote{\hyperlink{https://github.com/goodatlas/zeroth}{https://github.com/goodatlas/zeroth}}. To normalize Korean, we removed all punctuation (including brackets) and converted Western Arabic numbers to their phonetic spellings. For Japanese, we followed the neologism dictionary for MeCab~\footnote{\hyperlink{https://github.com/neologd/mecab-ipadic-neologd}{https://github.com/neologd/mecab-ipadic-neologd}}. Specifically, we leverage the default normalizer~\footnote{\hyperlink{https://github.com/ikegami-yukino/neologdn}{https://github.com/ikegami-yukino/neologdn}}. For Ukranian, Vietnamese, and Chinese, we leverage the basic text normalizer provided by Whisper.

\section{Complete Results}
\label{sec:exhaustive}

\input{figures/tables/all_results_tables.tex}

%% file: figures/tables/all_results_tables.tex
    \begin{table}[ht]
        \centering
    \begin{tabular}{lrrrr|rrrr}
\toprule
& \multicolumn{4}{c}{WER} & \multicolumn{4}{c}{CER} \\ \midrule
 Model                   &   CV17 &   Casablanca &   Fleurs &   SADA22 &   CV17 &   Casablanca &   Fleurs &   SADA22 \\
\midrule
 whisper-medium      &      36.09 &            76.2  &        18.79 &        69.48 &      12.4  &            39.16 &         5.82 &        44.39 \\
 whisper-small       &      48.02 &            87.22 &        29.09 &        88.55 &      17.39 &            46.19 &         9.1  &        57.08 \\
 whisper-base        &      80.35 &           108.33 &        50.15 &       111.39 &      38.21 &            63.16 &        17.7  &        73.92 \\
 whisper-tiny        &      88.89 &           109.32 &        66.01 &       109.76 &      44.02 &            66.98 &        25.7  &        72.93 \\
 moonshine-tiny (Baseline) &      77.54 &            97.8  &        79.16 &       100.82 &      40.41 &            60.75 &        40.97 &        61.5  \\
 moonshine-tiny (Final)    &      36.55 &            80.62 &        20.76 &        66.55 &      12.94 &            43.1  &         7.62 &        35.38 \\
\bottomrule
\end{tabular}
        \caption{Arabic}
        \label{tab:arabic-full}
    \end{table}

\begin{table}[ht]
        \centering
    \begin{tabular}{lrr}
\toprule
& \multicolumn{2}{c}{CER} \\ \midrule
 Model              &   CV17 &   Fleurs \\
\midrule
 whisper-medium      & 28.06 &        52.75 \\
 whisper-small       & 35.46 &        58.06 \\
 whisper-base        & 54.11 &        64.14 \\
 whisper-tiny        & 65.92 &        71.1  \\
 moonshine-tiny (Baseline) & 22.17 &        52.13 \\
 moonshine-tiny (Final)    & 36.1  &        29.44 \\
\bottomrule
\end{tabular}
        \caption{Chinese}
        \label{tab:chinese-full}
    \end{table}

\begin{table}[ht]
        \centering
    \begin{tabular}{lrrr}
\toprule
& \multicolumn{3}{c}{CER} \\ \midrule
 Model              &   CV17 &   Fleurs &   Reazon Speech \\
\midrule
 whisper-medium      &        29.09 &        11.5  &               43.06 \\
 whisper-small       &        37.69 &        16.76 &               68.37 \\
 whisper-base        &        71.13 &        27.13 &              119.81 \\
 whisper-tiny        &        96.11 &        47.2  &              146.81 \\
 moonshine-tiny (Baseline) &        30.99 &        19.16 &               11.28 \\
 moonshine-tiny (Final)    &        18.3  &        17.87 &               10.89 \\
\bottomrule
\end{tabular}
        \caption{Japanese}
        \label{tab:japanese-full}
    \end{table}

    \begin{table}[ht]
        \centering
    \begin{tabular}{lrrr|rrr}
\toprule
& \multicolumn{3}{c}{WER} & \multicolumn{3}{c}{CER} \\ \midrule
 Model              &   CV17 &   Fleurs &   Zeroth &   CV17 &   Fleurs &   Zeroth \\
\midrule
 whisper-medium      &      26.93 &        16.18 &        21.83 &       9.38 &         6.99 &         6.66 \\
 whisper-small       &      33.38 &        19.56 &        27.95 &      12.32 &         8.47 &         8.82 \\
 whisper-base        &      45.45 &        27.16 &        43.56 &      19.42 &        11.95 &        16.42 \\
 whisper-tiny        &      63.79 &        35.44 &        47.84 &      37.27 &        15.83 &        18.66 \\
 moonshine-tiny (Baseline) &      55.46 &        29.1  &        12.48 &      28.89 &        13.52 &         5.72 \\
 moonshine-tiny (Final)    &      35.22 &        20    &        16.38 &      14.94 &         8.9  &         5.72 \\
\bottomrule
\end{tabular}
        \caption{Korean}
        \label{tab:korean-full}
    \end{table}

\begin{table}[ht]
        \centering
    \begin{tabular}{lrr|rr}
\toprule
& \multicolumn{2}{c}{WER} & \multicolumn{2}{c}{CER} \\ \midrule
 Model              &   CV17 &   Fleurs &   CV17 &   Fleurs \\
\midrule
 whisper-medium      &      23.54 &        13.45 &      12.28 &         6.78 \\
 whisper-small       &      30.31 &        22.61 &      15.19 &        11.09 \\
 whisper-base        &      59.54 &        46.03 &      35.54 &        22.79 \\
 whisper-tiny        &     100.9  &        91.89 &      72.46 &        61.18 \\
 moonshine-tiny (Baseline) &     149.15 &        97.33 &     115.51 &        82.25 \\
 moonshine-tiny (Final)    &      18.83 &        13.01 &      10.26 &         6.74 \\
\bottomrule
\end{tabular}
        \caption{Vietnamese}
        \label{tab:vietnamese-full}
    \end{table}

    \begin{table}[ht]
        \centering
    \begin{tabular}{lrrr|rrr}
\toprule
& \multicolumn{3}{c}{WER} & \multicolumn{3}{c}{CER} \\ \midrule
 Model              &   CV17 &   Eurospeech &   Fleurs &   CV17 &   Eurospeech &   Fleurs \\
\midrule
 whisper-medium      &      20.9  &            17.01 &        11.62 &       5.93 &             8.99 &         3.68 \\
 whisper-small       &      32.44 &            24.94 &        20.41 &       9.03 &            11.1  &         5.33 \\
 whisper-base        &      55.55 &            46.06 &        44.08 &      18.4  &            18.6  &        14.51 \\
 whisper-tiny        &      67.07 &            69.41 &        63.83 &      24.03 &            28.93 &        20.69 \\
 moonshine-tiny (Baseline) &      92.49 &            64.93 &        92.28 &      63.49 &            44.21 &        68.12 \\
 moonshine-tiny (Final)    &      26.11 &            14.73 &        18.25 &       8.57 &             7.81 &         6.43 \\
\bottomrule
\end{tabular}
        \caption{Ukrainian}
        \label{tab:ukrainian-full}
    \end{table}